\documentclass[sigconf,nonacm]{acmart}

\usepackage{subfiles} 
\usepackage[linesnumbered,ruled]{algorithm2e}
\usepackage{multirow}
\usepackage{tabularx}
\usepackage{subcaption}
\usepackage{enumitem}
\usepackage{float}

\makeatletter
\newcommand{\removelatexerror}{\let\@latex@error\@gobble}
\makeatother

\AtBeginDocument{%
  \providecommand\BibTeX{{%
    \normalfont B\kern-0.5em{\scshape i\kern-0.25em b}\kern-0.8em\TeX}}}

\setcopyright{acmlicensed}
\copyrightyear{2024}
\acmYear{2024}

\acmConference[KDD '24]{30th SIGKDD Conference on Knowledge Discovery and Data Mining}{August 25--29, 2024}{Barcelona, Spain}
\begin{document}

\title{Personalised Drug Identifier for Cancer Treatment\\ with Transformers using Auxiliary Information}


\author{Aishwarya Jayagopal}
\affiliation{%
  \institution{National University of Singapore}
  \country{Singapore}
  }
\email{aishwarya.jayagopal@u.nus.edu}

\author{Hansheng Xue}
\affiliation{%
  \institution{National University of Singapore}
  \country{Singapore}
  }
\email{hansheng.xue@nus.edu.sg}

\author{Ziyang He}
\affiliation{%
  \institution{National University of Singapore}
  \country{Singapore}
  }
\email{heziyang@u.nus.edu}

\author{Robert J. Walsh}
\affiliation{%
  \institution{National University Cancer Institute}
  \country{Singapore}
  }
\email{robert_walsh@nuhs.edu.sg}

\author{Krishna Kumar Hariprasannan}
\affiliation{%
  \institution{National University of Singapore}
  \country{Singapore}
  }
\email{krishnakh@u.nus.edu}

\author{David Shao Peng Tan}
\affiliation{%
  \institution{Cancer Science Institute of Singapore}
  \country{Singapore}
  }
\email{david_sp_tan@nuhs.edu.sg}

\author{Tuan Zea Tan}
\affiliation{%
  \institution{Cancer Science Institute of Singapore}
  \country{Singapore}
  }
\email{csittz@nus.edu.sg}

\author{Jason J. Pitt}
\affiliation{%
  \institution{Cancer Science Institute of Singapore}
  \country{Singapore}
  }
\email{jason.j.pitt@nus.edu.sg}

\author{Anand D. Jeyasekharan}
\affiliation{%
  \institution{Cancer Science Institute of Singapore}
  \country{Singapore}
  }
\email{csiadj@nus.edu.sg}

\author{Vaibhav Rajan}
\affiliation{%
  \institution{National University of Singapore}
  \country{Singapore}
  }
\email{vaibhav.rajan@nus.edu.sg}







\renewcommand{\shortauthors}{Jayagopal and Xue, et al.}

\begin{abstract}
Cancer remains a global challenge due to its growing clinical and economic burden.
Its uniquely personal manifestation, which makes treatment difficult, has fuelled the quest for personalized treatment strategies.
Thus, genomic profiling is increasingly becoming part of clinical diagnostic panels.
Effective use of such panels requires accurate
drug response prediction (DRP) models, which are challenging to build due to limited labelled patient data.
Previous methods to address this problem have used 
various forms of transfer learning.
However, they do not explicitly model the variable length sequential structure of the list of mutations in such diagnostic panels.
Further, they do not utilize 
auxiliary information (like patient survival) for model training.
We address these limitations 
through a novel transformer-based method, 
which surpasses the performance of state-of-the-art DRP models on benchmark data.
Code for our method is available at \href{https://github.com/CDAL-SOC/PREDICT-AI}{https://github.com/CDAL-SOC/PREDICT-AI}.

We also present the design of a treatment recommendation system (TRS), which is currently deployed at the National University Hospital, Singapore and is being evaluated in a clinical trial.
We discuss why the recommended drugs and their predicted scores alone, obtained from DRP models, are insufficient for treatment planning.
Treatment planning for complex cancer cases, in the face of limited clinical validation, requires assessment of many other factors, including several indirect sources of evidence on drug efficacy.
We discuss key lessons learnt on
model validation and use of indirect supporting evidence to build clinicians’ trust and aid their decision making.

\end{abstract}





\maketitle



\section{Introduction}
\begin{figure*}[t]
\centering
\includegraphics[width=1.75\columnwidth]
{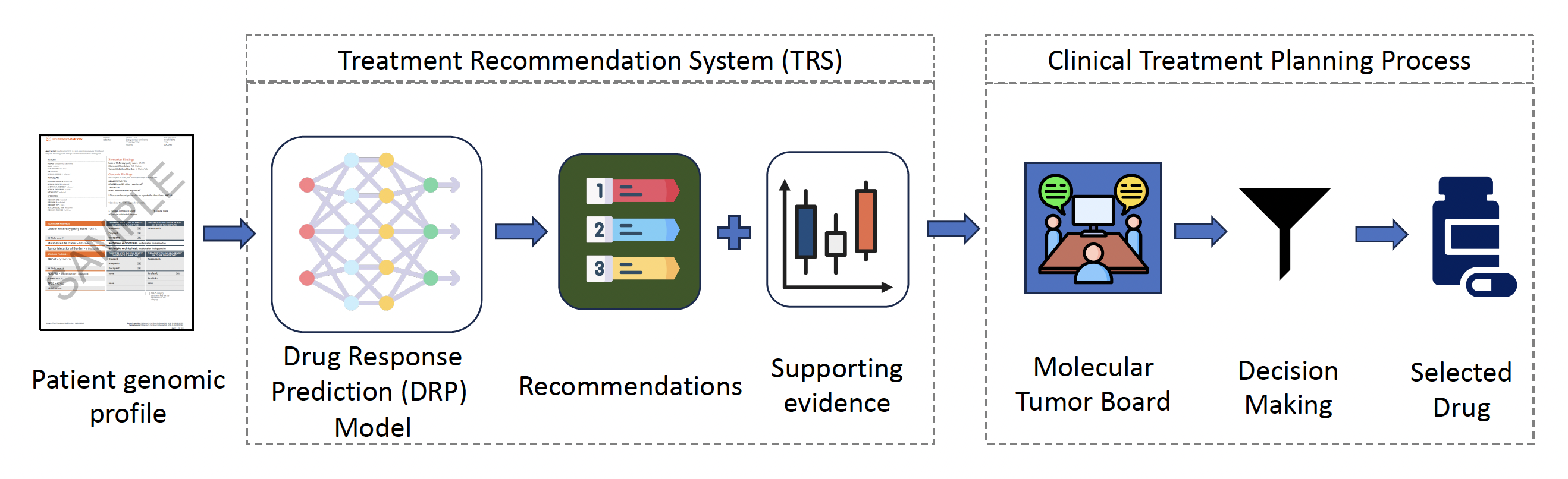}
  \caption{Overview of our personalized Treatment Recommendation System (TRS) and its role in clinical treatment planning.}
  \label{fig:teaser}
\end{figure*}
Cancer, a leading cause of death globally, occurs due to the uncontrolled proliferation of cells. During cancer progression, cells in the body acquire specific alterations (\textit{mutations}) that allow them to multiply rapidly compared to normal cells, resulting in tumors. 
Cancer remains difficult to cure due to its high complexity, which is not yet fully understood, and considerable heterogeneity in treatment outcomes~\cite{wahida2023coming}.
While clinical trials have identified therapeutic agents that target specific mutations, these studies have largely been restricted to mutations in individual genes~\cite{brachova2013consequence, randic2023single}. 
Conducting trials to identify the most suitable drug for every combination of mutations across all ($\sim20000$) genes is intractable due to the exponentially large combinatorial space of mutations. Machine learning solutions provide a promising direction and can 
mine patterns from patient genomic profiles that correlate with drug response. However, building these drug response prediction (DRP) models require large patient genomic datasets with documented response to different drugs. While patient genomic profiles have been collected in large numbers~\cite{cerami2012cbio, weinstein2013cancer}, corresponding drug response data is limited as standard of care treatment~\cite{van2010advanced, planchard2018metastatic} in clinics only involve a narrow subset of drugs decided by the cancer type.

This dearth of patient drug response data has led researchers to create and explore  `proxy' patient datasets -- \textit{cell lines}. These are cancer cells extracted from patients, which can be cloned (thus having the same genomic profiles)
and simultaneously screened for multiple drugs~\cite{depmap}.
These cells proliferate
in an environment different from 
the human body, and this results in distributional shifts in the mutations and changes in the observed drug responses, compared to those in patients. 
Thus, DRP models on cell lines are not directly usable on patients. 
To address this problem, several transfer learning methods, e.g., ~\cite{he2022context, sharifi2021out}, have been developed to effectively utilize both the labelled 
source domain (cell lines) and 
target domain (patients) containing some labelled and relatively abundant unlabelled data. These methods have not been evaluated in a clinical setting which pose additional data-related challenges.

Omics data is rapidly becoming part of clinical diagnostics for cancer.
Patients undergo genomic sequencing to identify mutations present in a \textit{subset} of cancer-related genes. For example, popular clinical sequencing panels like FoundationOne CDx~\cite{milbury2022clinical} and TruSight Oncology 500~\cite{wei2022evaluation} consider 324 and 523 genes respectively. 
Note that each gene is a sequence of DNA alphabets and a mutation can occur in any location in the sequence.
The sequencing report lists out mutations detected in the genes (at a sub-gene level) and their locations -- the effect of a mutation depends on the location as well. 
Thus, input data for a patient comprises a sequence of mutations, with the sequence length varying  across patients. 

Previous transfer learning based DRP models have several limitations:
(i) The best previous methods~\cite{he2022context, sharifi2021out} have considered subsets of genes (e.g., 1426 in CODE-AE, 2128 in Velodrome) based on criteria like maximum variations in values and known interactions amongst proteins. The selected subsets do not match those in diagnostic panels.
(ii) Most prior methods, e.g.,~\cite{he2022context, sharifi2020aitl, sharifi2021out, ma2021few, peres2021tugda}, consider gene expression data (indicative of gene activity) to train DRP models, which is largely unavailable in current clinical settings~\cite{zehir2017mutational, kou2016possibility}. 
(iii) To our knowledge, DruID~\cite{jayagopal2023multi} is the only DRP model that addresses the above problems. This work also shows that to  effectively utilize diagnostic panels, sparsity of the mutation data (a problem which previous gene expression based models do not face) has to be carefully modeled. Further, DruID also used the sub-gene level variant information, through additional annotations, which predict the effect of the mutation locations on cancer. However, these annotations were aggregated to obtain a fixed-dimensional representation of each gene which was subsequently used as input to the DRP model.
Thus, none of the extant methods explicitly model the variable-length sequential nature of the inputs.
(iv) The best previous methods, including DruID, follow the paradigm of unsupervised representation learning (to utilize unlabelled data) followed by supervised fine tuning. They have 
largely neglected available auxiliary information which are highly correlated to drug response, such as patient survival, which can be used to supervise representation learning. 

We address these limitations through a novel transformer-based algorithm \textbf{PREDICT-AI} (PeRsonalisEd Drug Identifier for Cancer Treatment using Auxiliary Information), to predict drug response using sparse mutation data from diagnostic panels. 
Following~\cite{jayagopal2023multi}, we annotate sub-gene level mutation locations, to model their functional effects, and obtain a feature vector per input mutation.
We design a novel tokenization approach to embed this sequence of feature vectors
in transformers.
Our model learns from both cell line and patient mutation data, and utilizes auxiliary survival information,
in a two-stage training process comprising: (i) Transformer enhanced Multi-Task Logistic Regression-based Survival Prediction \texttt{TransformerMTLR}, and (ii) Pre-trained Transformer-based Drug Response Prediction \texttt{TransformerDRP}. We evaluate the performance of PREDICT-AI on benchmark datasets and show that our modeling approach leads to substantial performance improvement over state-of-the-art methods.

Although several DRP methods have been developed in the past, their application in mainstream clinical practice is limited~\cite{mittermaier2023collaborative, corti2023artificial}.
Using a DRP model on a set of drugs, we can predict which drug is likely to be most effective for a patient. However, this prediction alone is insufficient for clinicians to decide whether or not to administer the drug, which also makes clinical validation of a DRP model challenging. Treatment planning for complex cancer cases is increasingly done by a Molecular Tumor Board (MTB)~\cite{tsimberidou2023molecular, luchini2020molecular}, where several expert clinicians collectively decide on the most suitable treatment. Their decision-making relies on knowledge from clinical trials and/or mechanism of action for each drug as well as clinical guidelines. However, mechanism of action of drugs may not always be fully known.
And, as mentioned earlier, it is infeasible to obtain knowledge of drug efficacy for every combination of mutations through clinical trials. 
As a result, the choice of a drug for a patient, has to be made by considering multiple incomplete or indirect sources of evidence. In fact, this would be the decision-making process even to evaluate the efficacy of a DRP model through a clinical trial.
We discuss these considerations, steps taken towards building such a treatment recommendation system (TRS) for MTB (Figure \ref{fig:teaser}) and lessons learnt from the deployment.

To summarize, our main contributions are as follows:
\begin{itemize}[noitemsep,topsep=0pt,labelindent=0em,leftmargin=*]
    \item We design a new transformer-based machine learning method, \texttt{PREDICT-AI}, which predicts the efficacy of a given drug on a genomic profile comprising mutation information of a
    subset of genes sequenced in clinical diagnostic reports. 
    \item To the best of our knowledge, we are the first to incorporate auxiliary information, like patient survival, indicative of patient response in transfer-learning based drug response prediction. 
    \item We propose a novel tokenization approach to explicitly model variable length sequential inputs and incorporate biologically meaningful annotations of individual mutations as features.
    \item We describe our implementation of a treatment recommendation system (TRS), hosted at \href{https://pharmacope.ai/}{https://pharmacope.ai/}, which uses DRP models, to support molecular tumor boards (MTB) in their treatment planning process. We discuss key lessons learnt on model validation and use of indirect supporting evidence to build
    clinicians' trust and aid their decision making.
\end{itemize}

\section{Background and Related Work}

\subsection{Background}
\label{sec: bkgd}
Dearth of drug response data in patients has led researchers to create and explore 
\textit{cell lines}.
These are cancer cells extracted from patients, which are replicated and studied under controlled laboratory conditions~\cite{depmap}. Replicated cells, having identical genomes, are subjected to different doses of various drugs to examine their response. A common measure of drug response in cell lines is the \textbf{Area Under the Dose Response Curve (AUDRC)}, which is a real-valued function of the number of surviving cells after exposure to the drug. 
These cell lines are in an environment different from and outside the patient body, and so, mutations observed across cancer cells in cell lines and patients display distributional differences. Further, their response to drugs may also be different from that of a tumour inside a patient. 

Patient drug response is measured as a function of change in tumor volume, using \textbf{Response Evaluation Criteria In Solid Tumors (RECIST)}~\cite{therasse2000new} categories - Stable Disease, Progressive Disease, Complete Response, Partial Response. The four categories are coalesced into 2 classes indicating good/bad response.
Some patient datasets also measure the time taken from treatment start to tumor progression/death, i.e. \textbf{Progression-Free Survival (PFS)}~\cite{buyse2007progression}, which can be indicative of drug efficacy. 
DRP
models trained on cell lines cannot translate directly to patients~\cite{sharifi2020aitl, mourragui2019precise} due to distributional differences in mutations, differences in drug response measurements and functions. Thus, transfer learning approaches are used to effectively utilize data from both domains. 

\subsection{Related Work}
Prior DRP models perform transfer learning between the source domain (cell lines) and target domain (patients). These methods can be categorised broadly based on their use of labeled patient samples. Inductive transfer learning methods, like AITL~\cite{sharifi2020aitl} and TCRP~\cite{ma2021few} used both labeled cell line and patient samples. Transductive methods like TUGDA~\cite{peres2021tugda}, Velodrome~\cite{sharifi2021out} used labeled cell line and unlabeled patient samples. Only a few methods, like CODE-AE~\cite{he2022context} and DruID~\cite{jayagopal2023multi}, relied on unsupervised transfer learning using available unlabeled cell line and patient datasets in a pre-training stage. Most of these methods derived binary patient labels from RECIST categories and none of them have used PFS information for DRP modeling. 
All these methods (except DruID) used gene expression data, and have not been evaluated on sparse mutation data from diagnostic panels. 
DruID used mutation level variant annotations to model their functional effects but has not explictly modeled 
their variable-length sequential structure. Through the used of transformers, our model addresses this limitation and achieves superior empirical performance on benchmark data.

Many survival models have been developed to predict time-to-event \cite{lee2019review,wang2019machine}, the event being tumor progression in our case. Here, we mention a few recent works, particularly those based on deep learning. These methods can be broadly classified into two types based on whether they model survival time as discrete or continuous. Models like DeepSurv~\cite{katzman2018deepsurv}, PCHazard~\cite{kvamme2019pchazard} treat survival time as continuous, while others like LogisticHazard~\cite{gensheimer2019logistichazard}, PMF, DeepHit~\cite{lee2018deephit} and MTLR~\cite{fotso2018mtlr} treat it as discrete. These methods largely rely on simpler neural networks and empirically yield poor performance on relatively small datasets. Our approach in this work is based on MTLR but we use
transformers in place of feed forward networks and integrate survival prediction with DRP.


\section{Data}
We obtain mutation profiles of cell lines from the Cancer Cell Line Encyclopedia (CCLE) \href{https://depmap.org/portal/download/all/?releasename=DepMap+Public+21Q3}{DepMap portal}~\cite{depmap} and their drug responses from the \href{https://www.cancerrxgene.org/downloads/bulk_download}{Genomics of Drug Sensitivity in Cancer (GDSC) portal}~\cite{yang2012genomics}. The mutation data for patients with RECIST response, was obtained from \href{https://portal.gdc.cancer.gov/}{The Cancer Genome Atlas (TCGA) GDC portal}~\cite{weinstein2013cancer}. The corresponding RECIST response data was obtained from~\cite{jia2021deep} and converted to binary labels as in~\cite{peres2021tugda}. There is a significant class imbalance in this dataset, with 454 positively and 175 negatively labeled (sample, drug) pairs. For patients with survival information, we obtained both mutation profiles as well as progression-free survival information for colorectal cancer patients from the  \href{https://genie.cbioportal.org/study/summary?id=crc_public_genie_bpc}{GENIE BPC CRC v2.0-public dataset}~\cite{wang2023evaluation} (henceforth `\textbf{CRC}') and for non-small cell lung cancer patients from the \href{https://genie.cbioportal.org/study/summary?id=nsclc_public_genie_bpc}{GENIE BPC NSCLC v2.0-public dataset}~\cite{choudhury2023genie} (henceforth `\textbf{NSCLC}'). Drugs were encoded using Morgan fingerprints~\cite{morgan1965generation}. Details of processing are in Appendix \ref{sec:data_processing}. Data statistics after processing are in Table \ref{tab:data_stats}.

\begin{table}[!ht]
  \caption{Datasets used in patient and cell line domains with number of processed samples and drugs. }
  \label{tab:data_stats}
  \begin{tabular}{p{1.2cm}p{0.8cm}p{1.5cm}p{1.5cm}p{1.5cm}}
    \toprule
    Domain & Dataset Name & Response variable & Processed samples & (Sample, Drug) pairs \\
    \midrule
    Cell lines & CCLE & AUDRC & 689 & 3632 \\
    \midrule
    Patients & TCGA & RECIST & 470 & 629 \\
    & CRC & PFS & 822 & 822 \\
    & NSCLC & PFS & 1490 & 14910 \\
    \bottomrule
  \end{tabular}
\end{table}


\section{Method}
\subsection{Preliminaries}
The patient records utilized for drug response prediction tasks primarily include information on genes, mutations, and drugs. In our experiments, we consider 324 genes (from the FoundationOne CDx panel) associated with patients, denoted as $G$, 23-dimensional annotation-based features for each 
mutation, denoted as $M\in\mathbb{R}^{|M|\times 23}$, and drugs with Morgan fingerprint features represented by $D\in\mathbb{R}^{|D|\times 2048}$. 
We consider 3 separate tasks, each with its own dataset. i) For the survival prediction task, the training dataset is a combination of the CRC and NSCLC datasets, denoted as $\mathcal{P}=\mathcal{P}_{CRC}\cup\mathcal{P}_{NSCLC}$. These patient samples not only include three types of information (genes, mutations, and drugs) but also encompass survival time information ($t$), denoted as $\mathcal{P}=\{g_{ui},m_{ui},d_i,t_i\}_{i=1}^{N}$, where $g_{ui}\subseteq G$, $m_{ui}\subseteq M$, $d_i\subseteq D$, $t_i\in\mathbb{R}$, $N$ denotes number of patients and $u \in \mathbb{Z}^{+}$ varies across patients. 
ii) In the drug response prediction task, TCGA patient data is employed and denoted as $\mathcal{P}_{TCGA}$. Drug response prediction aims to acquire information about patient samples, specifically genes and mutations, to predict their drug responses as a classification task. Similar to i), the TCGA patient dataset can be represented as $\mathcal{P}_{TCGA}=\{g_{ui},m_{ui},d_i,y_i\}_{i=1}^{N}$, where $y_i \in \{0, 1\}$ where $y_i = 1$ indicates that the patient responds well to $d_i$ and $y_i = 0$ indicates that the tumor worsens. 
iii) Additionally, during response prediction, we integrate the task for predicting the real-valued AUDRC on cell lines, to improve the overall performance of response prediction~\cite{jayagopal2023multi, sharifi2020aitl}. The cell line data is employed in this task, encompassing information on genes, mutations, drugs, and associated AUDRC values, denoted as $\mathcal{P}_{CL}=\{g_{ui},m_{ui},d_i,v_i\}_{i=1}^{N}$, where $v_i \in [0, 1]$ with lower value indicating better response to the drug. It must be noted that in all 3 cases, the number of gene-mutation pairs $u$ can differ across samples, resulting in a variable length input. 

\begin{figure*}[t]
\centering
\includegraphics[width=1.85\columnwidth]{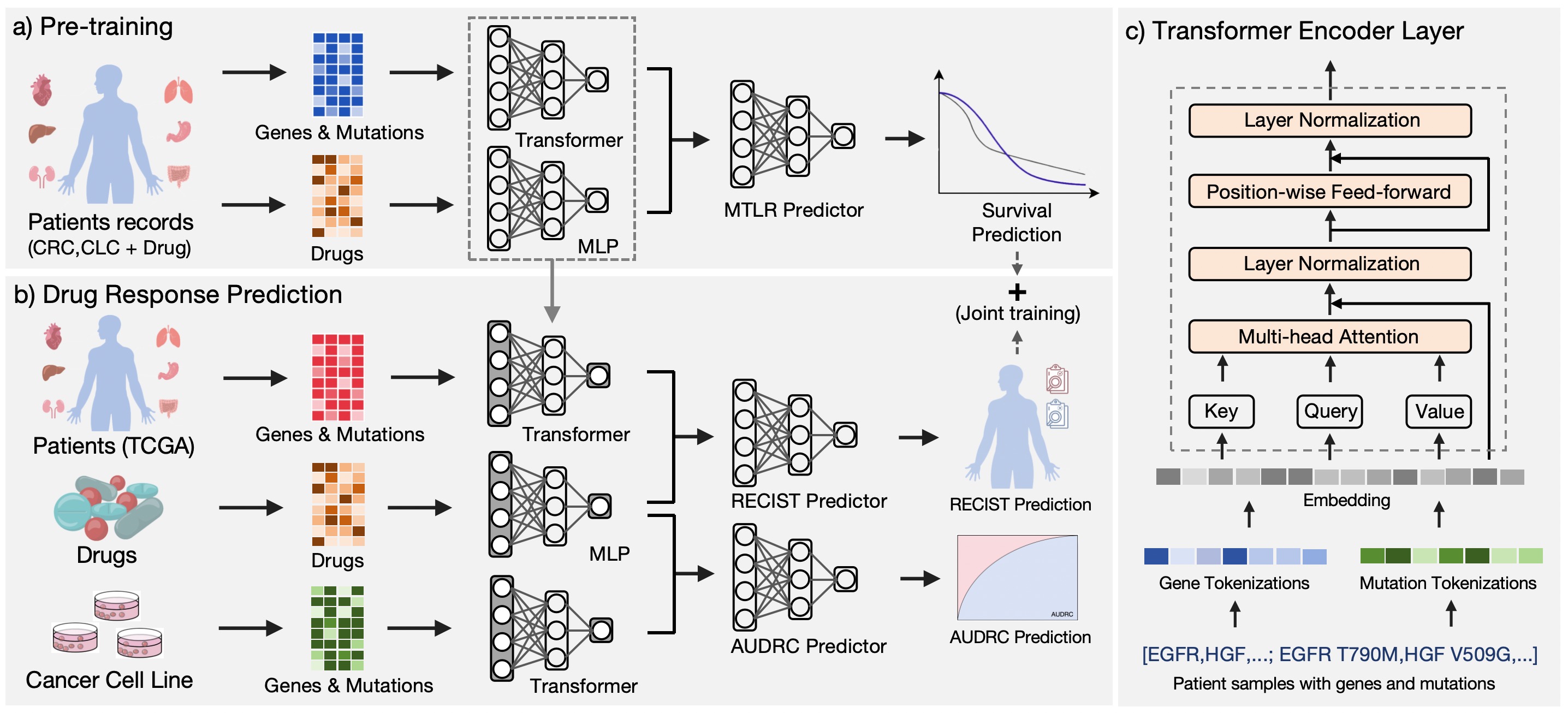}
\caption{Overview of the \texttt{PREDICT-AI}, consisting of three main components: (A) \texttt{TransformerMTLR}: a multi-task logistic regression model based on transformers for survival prediction. (B) \texttt{TransformerDRP}: a pretrained transformer-based drug response prediction model incorporating the AUDRC prediction task. (C) Detailed description of the transformer encoder layers.}
\label{fig01}
\end{figure*}

In summary, our model involves three tasks: survival prediction formulated as multi-task logistic regression, AUDRC prediction as a regression task, and the RECIST prediction as a classification task. In the initial pre-training step, we perform survival prediction using CRC and NSCLC data. Subsequently, we leverage both survival prediction on NSCLC data and AUDRC prediction on cell line data to improve the overall performance of the final drug response prediction in TCGA patients. 
The $\texttt{PREDICT-AI}$ model primarily comprises two components: a multi-task logistic regression model based on transformers for survival prediction (\texttt{TransformerMTLR}) and a pretrained transformer-based drug response prediction model incorporating the AUDRC prediction task (\texttt{TransformerDRP}). The overall architecture is illustrated in Figure~\ref{fig01}.

\subsection{Tokenization}
\label{sec: tokenization}
\begin{figure}[!ht]
\centering
\includegraphics[width=\linewidth]{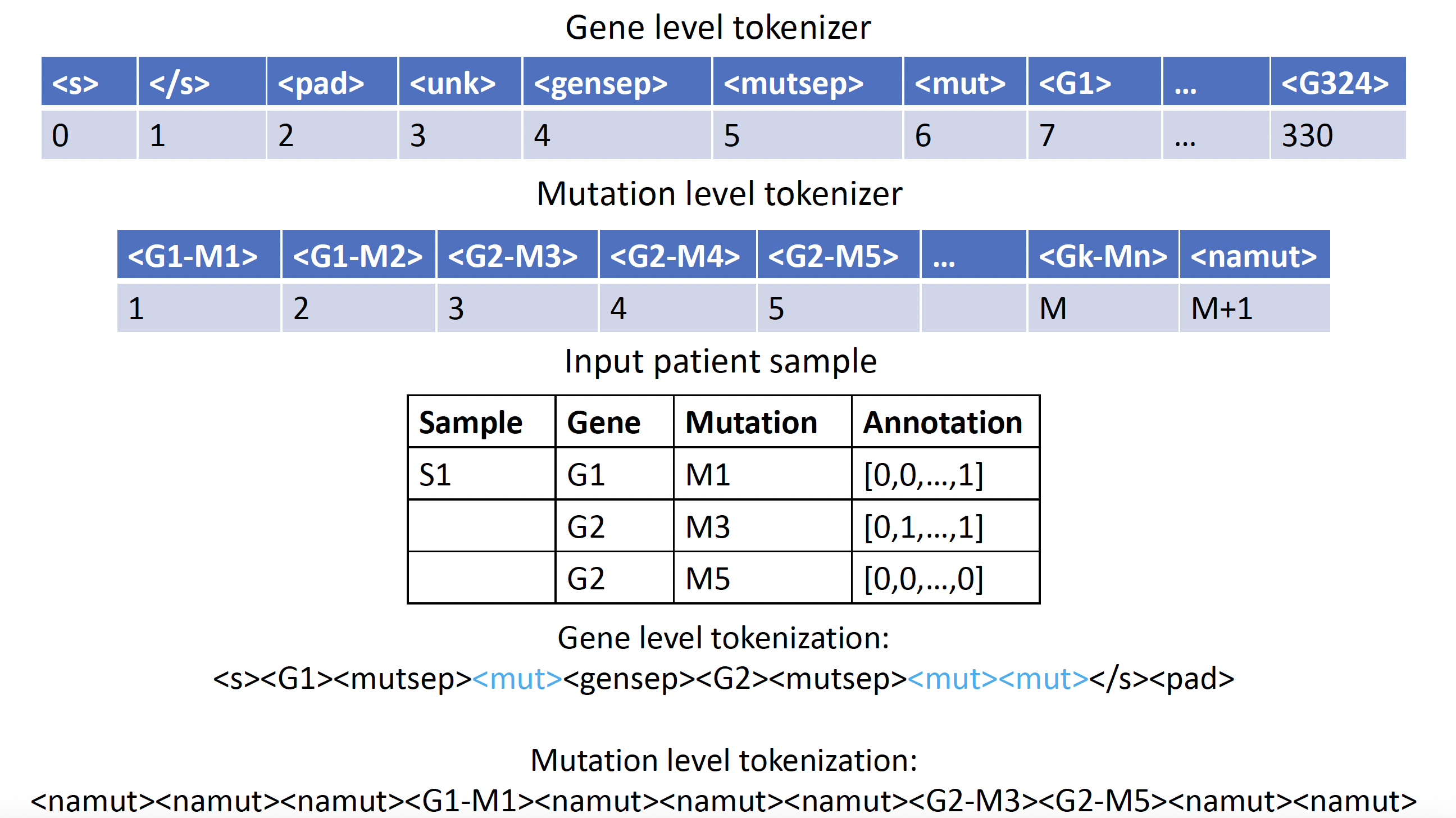}
\caption{Overview of proposed tokenization procedure with gene and mutation tokenizers.}
\label{fig_token}
\end{figure}

We use transformers to model varying length inputs, with each input being a sequence of 
\{\texttt{gene-mutation}\} names such as \texttt{\{TP53-R306\}, \{KRAS-G12V\}},... 
Note that there can be multiple mutations in the same gene and the number of listed genes can differ across patients.
The first step is to tokenize the inputs without treating the names of genes and mutations as linguistic words.
To this end, we propose a novel tokenizer 
as shown in Figure \ref{fig_token}.

We have two separate tokenizers - one each at the gene and mutation levels. The gene-level tokenizer vocabulary has 324 gene-specific tokens (1 per gene), 2 special separator tokens (\texttt{<gensep>}, \texttt{<mutsep>}), 4 classical special tokens (\texttt{<s>, </s>, <pad>, <unk>}), and 1 general mutation token (\texttt{<mut>}).  As shown in Figure \ref{fig_token}, each of these tokens is mapped to a unique index ranging from 0 to 330. \texttt{</s>} and \texttt{<s>} denote start and end of sequence. \texttt{<unk>} indicates unknown tokens and \texttt{<pad>} is for padding shorter sequences. \texttt{<gensep>} is used to separate out genes. \texttt{<mut>} is used to denote the presence of a mutation within the gene.
In Figure \ref{fig_token}, given a sample S1, the gene level tokenization includes the 4 classical special tokens. Genes G1 and G2 are assigned the corresponding gene-specific token with \texttt{<mutsep>} used to denote start of mutations within a gene (e.g., before M3, M5 in G2). Each individual mutation is assigned the token \texttt{<mut>}.

In the mutation level tokenizer, each \texttt{gene-mutation} pair is mapped to a unique number, ranging from 1 to the maximum number of \texttt{gene-mutation} pairs ($M$) (Figure \ref{fig_token}). Each token is associated with a set of 23 annotated features $M \in \mathbb{R}^{|M| \times 23}$. Details of the annotations can be found in ~\cite{wang2010annovar, landrum2018clinvar, li2020protein}.
Tokens that are not recognised as \texttt{gene-mutation} pairs in the genomic profile are mapped to index $M+1$ and token \texttt{<namut>}, and embedded as a 23-dim unit vector $\mathbf{1}^{23}$. The tokenized output lengths of both tokenizers is identical. 

Given an input sample, the final vector representation, in terms of tokens, is made up of two parts - one each from the gene and mutation level tokenizers. A gene-level vector is made up of gene level token indices, extracted from the gene tokenizer, and padded to reach maximum possible gene index. Similarly a mutation level vector is made up of mutation level token indices and padded to reach maximum possible gene-mutation index. These two vectors are concatenated and passed as input to the transformer.

\subsection{\texttt{TransformerMTLR}: Multi-Task Logistic Regression- based Survival Prediction}
The survival prediction model primarily comprises two components: i) formulating survival prediction as a multi-task logistic regression model and ii) constructing transformer and neural network models to learn from patients' clinical inputs.\\ 

\noindent\textbf{i. Multi-task logistic regression.} 
We formulate the survival prediction as a multi-task logistic regression task following \citep{fotso2018mtlr}. 
For a patient $p\in\mathcal{P}$ with survival time $t$, the survival function can be computed as $\mathcal{F}(t) = Prob(t'>t)$, representing the probability that an individual within the population will survive beyond time $t$~\citep{yu2011learning}. 
When provided with the survival times of a group of individuals, we can graphically depict the proportion of individuals surviving over time 
using the Kaplan-Meier curve, which is closely associated with the hazard function. 
Therefore, the survival function can be computed as:
$\mathcal{F}(t)=\exp(-\int_{0}^{t}\lambda(u)du)$, where $\lambda(t)=\lim_{\Delta t\to 0}\frac{P(t\le t'< t+\Delta t|t'\ge t)}{\Delta t}$.
Many methods exist for estimating $\mathcal{F}(t)$ and $\lambda(t)$, such as Cox's proportional hazards model~\citep{cox1972regression} and the Kaplan-Meier model~\citep{kaplan1958nonparametric}. However, these models are computationally inefficient and rely on strict distributional assumptions. 

Multi-task logistic regression can mitigate these limitations by utilizing a series of logistic regression models across various time intervals to estimate the probability of the event of interest occurring within each interval. Thus, the response variable $s_j$ for a logistic regression model on each interval $\alpha_j$ can be formulated as $s_i=1$ if $t'\in\alpha_j$ i.e., the event happened in the interval $\alpha_j$ else 0. 
Thus, if a patient experiences an event (tumor progression in this case) in one of $k$ intervals ($k \in [1, K]$), the response vector with $K$ time intervals can be formulated as $S=[s_0=0,...,s_{k-1}=0,s_k=1,...,s_K=1]$.
To capture the underlying non-linear relationships in the data, neural multi-task logistic regression was introduced, incorporating multi-layer perceptrons~\citep{fotso2018mtlr}.The survival function can be defined as:
\begin{equation}\label{eqn01}
    \mathcal{F}(t_{k-1},x)=\sum_{i=k}^{K}\frac{\exp{(\phi(x)\cdot\Delta\cdot S)}}{Z(\phi(x))},
\end{equation}
where $\phi:\mathbb{R}^{d}\longmapsto\mathbb{R}^{K}$ is the transformation function, $x\in\mathbb{R}^d$ denotes $d$-dimensional features. $\Delta\in\mathbb{R}^{K\times (K+1)}$ is a triangular matrix. $Z(\phi(x))=\sum_{j=1}^{K}\exp(\sum_{l=j+1}^{K}\phi(x))$ is the normalizing constant. 
For survival prediction, we define the loss function as the negative log-likelihood of Equation~\ref{eqn01} (similar to DeepSurv~\citep{katzman2018deepsurv}). The log-likelihood function primarily comprises two components—one for uncensored instances $\mathcal{P}_u$ and one for censored instances $\mathcal{P}_c$. The whole sample data is denoted as $\mathcal{P}=\mathcal{P}_u\cup\mathcal{P}_c=\{\bar{X},S\}$, where $\bar{X}$ denotes the learned features from transformer or drug embedder (described in the following sections). The objective function for survival prediction can be formulated as:
\begin{equation}
\begin{aligned}
 \mathcal{L}_{S} = & \sum_{j\in\mathcal{P}_u}\sum_{k=1}^{K-1}\phi_{k}(X^{j})S_{k}^{j}+
 \sum_{j\in\mathcal{P}_c}\log\sum_{t_k>T_c^{j}}\exp(\sum_{k=1}^{K-1}\phi_k(X^j))\\
 & -\sum_{j=1}^{|\mathcal{P}|}\log Z(\phi(X^j))
\end{aligned}
\end{equation}
\noindent\textbf{ii. Transformer and Neural networks.}
Within the multi-task logistic regression model, the learning module comprises a transformer for patients and a multilayer perceptron for drugs. Following the utilization of these learning modules, a transformation is employed for prediction, $h$. Subsequently, we will provide a detailed introduction to the transformer utilized for patient learning. 
%
Given patient samples $\mathcal{P}$, we perform tokenization and obtain the feature vector as in Section \ref{sec: tokenization}.
The features for each gene and mutation can be represented as $X_g \in \mathbb{R}^d$ and $X_m \in \mathbb{R}^{d}$, respectively. The combined features for each patient are denoted as $X=[X_g||X_m]$, 
where $||$ denotes the concatenation operation. 

The structure of the transformer encoder layer primarily includes one multi-head attention layer, one position-wise feed-forward layer, and several layer normalization operators~\cite{vaswani2017attention}. 
The multi-head attention layer trains multiple attention heads independently to focus on distinct parts of the inputted patients' embeddings. 
Given the input encoding $X$, three distinct representations known as query (Q), key (K), and value (V) are derived through a linear transformation of $X$, expressed as $Q = XW^Q$, $K = XW^K$, and $V = XW^V$, where $W^Q, W^K, W^V \in \mathbb{R}^{d\times d}$ represent trainable parameters of the network. The self-attention mechanism is defined as:
\begin{equation}
    \mathrm{Attention}(Q,K,V) = \mathrm{softmax}\left(\frac{QK^\top}{\sqrt{d}}\right)V.
\end{equation}
The attention mechanism uses the scaled dot product of the key and query to calculate the relevance between the sequence elements, and then the softmax function is applied on the $QK^\top$. The scaling factor $1/\sqrt{d}$ works analogously to standard dot product attention. The resulting matrix contains a set of weights in each row to compute the weighted sum of the input values $V$. To allow the model to attend to information based on varying patterns multiple attention-heads are employed. 
\begin{gather}
    \mathrm{MultiHead}(X)=\mathrm{Concat}(\mathrm{head}_1,\dots,\mathrm{head}_h)W^O,\\
    \mathrm{where}\ \mathrm{head}_i=\mathrm{Attention}(XW^Q_i,XW^K_i,XW^V_i).
\end{gather}
where $h$ represents the number of attention heads, and $W^Q$, $W^K$, $W^V$, and $W^O$ stand for trainable parameters of the network. Following the multi-head attention operation, layer normalization and residual connections are employed to facilitate training acceleration, enhance generalization, and alleviate the vanishing or exploding gradients problem. Additionally, a fully connected feed-forward network (FFN) is applied uniformly to each token. This layer comprises two linear transformations with a rectified linear unit (ReLU) activation in the middle.

In the survival prediction module, a neural network, such as a multi-layer perceptron, incorporating batch normalization and activation functions, is utilized to learn the features of the inputted fingerprints for drugs and serve as the predictor. The structure of the neural network can be expressed as: 
\begin{equation}
    \bar{X}_D = \mathrm{MLP}_{Drug}(X_D),\ \Hat{y}=\mathrm{MLP}(\bar{X}_D||\bar{X}),
\end{equation}
where $X_D$ is the 2048-dimensional fingerprint features of drugs, $\bar{X}$ is the learned features outputted from transformer, and $||$ denotes the concatenation operation.

\subsection{\texttt{TransformerDRP}: Pre-trained Transformer-based Response Prediction}

Within the \texttt{TransformerDRP}, we primarily incorporate two drug response prediction tasks -- based on patient data (RECIST predictor) and based on cell line data (AUDRC predictor).\\
\noindent\textbf{i. RECIST Predictor.} By optimizing the loss function in Equation \ref{eqn01} within the survival prediction section, we can acquire trained transformers for patients and neural networks for drugs. During this stage, we employ these pretrained models to predict drug response by jointly optimizing RECIST prediction and survival prediction. 
Similar to the patients (CRC and NSCLC) in the survival prediction section, patient samples (TCGA) employed in the response prediction part also encompass genes, mutations, and drugs for each sample. 
We input the patient and drug features into the transformer and neural network, respectively, to obtain the learned embeddings $\bar{X}_T$ and $\bar{X}_D$. Subsequently, a MLP-based RECIST predictor is constructed to predict the response, $\hat{Y}=\mathrm{MLP}(\bar{X}_T||\bar{X}_D)$. 
Next, to address the class imbalance problem, we utilize the BCEFocalLoss as the objective function, calculated as:
\begin{equation}
    \mathcal{L}_R = \frac{1}{N}\sum_{i\in N}[-\alpha y_i(1-\hat{y}_i)^{\gamma}\log(\hat{y}_i)-(1-\alpha)(1-y_i)\hat{y}_i^{\gamma}\log(1-\hat{y}_i)],
\end{equation}
where, $\hat{y}\in\hat{Y}$ represents the predicted response, and $y\in Y$ is the true response. $N$ denotes the number of patient samples. The weighting factor is denoted as $\alpha \in [0,1]$, and $\gamma \geq 0$ serves as the tunable focusing parameter.

\noindent\textbf{ii. AUDRC Predictor.} 
Note that the format of the preprocessed cell line data is similar to the patient samples. 
Through the pretrained transformer and drug neural network, we can also obtain the embeddings for samples and drugs $\hat{X}_C$ and $\hat{X}_D$. A MLP-based AUDRC predictor is constructed to predict the value, $\hat{R}=\mathrm{MLP}(\bar{X}_C||\bar{X}_D)$. 
The MSE loss function for the AUDRC predictor and calculated as:
\begin{equation}
\mathcal{L}_C = \frac{1}{M}\sum_i^M(\hat{r}_i-r_i)^2,
\end{equation}
where, $\hat{r}_i\in\hat{R}$ represents the predicted response, and $r\in R$ is the true AUC value. $M$ denotes the number of cell line samples. In the response prediction phase, we jointly optimize three objective functions, $L_S$ for survival prediction, $L_R$ for response prediction, and $L_C$ for AUDRC predictor. The pseudocode for \texttt{PREDICT-AI} is shown in Algorithm 1 \ref{algorithm}.


\section{Experiments and Results}
To evaluate our model, \texttt{PREDICT-AI}, we conduct 3 experiments - i) evaluation of \texttt{TransformerMTLR} on survival prediction, ii) evaluation of \texttt{PREDICT-AI} against state-of-the-art DRP models, iii) ablation tests on various model components. 

\subsection{Survival Prediction with \texttt{TransformerMTLR}}
To evaluate \texttt{TransformerMTLR}, we compare 
its test set concordance index \cite{schmid2016use} with that of 6 baseline survival models from the  \texttt{pycox} library \cite{pycox}. Each input for baseline models is a binary vector $\in \{0, 1\}^{324}$ representing the absence (0) or presence (1) of any mutation in a gene. All models are trained on same train set of NSCLC and CRC datasets, which is divided into train, validation, test splits (64:16:20) using same random seed in our experiment. As shown in Table \ref{tab:surv_results}, our method outperforms the second-best \texttt{DeepSurv} method by 5.9\%. 
We also explore the effect of using annotated mutations as inputs to these baseline models (Appendix Table \ref{tab:surv_results_2}). 

\begin{table}[!h]
  \caption{Comparison of TransformerMTLR against survival prediction baselines on NSCLC and CRC datasets.}
  \label{tab:surv_results}
  \begin{tabular}{p{1.6cm}p{1.5cm}|p{1.5cm}p{1.5cm}}
    \toprule
    Model & Test CI  & Model & Test CI \\
    \midrule
    DeepSurv & 0.5834 & MTLR & 0.5624 \\
    DeepHit & 0.4841  & PCHazard & 0.3197 \\
    LogisticHazard & 0.5121 & PMF & 0.5072 \\
    \midrule
    \multicolumn{4}{c}{TransformerMTLR \textbf{0.6425}}\\
    \bottomrule
  \end{tabular}
\end{table}

\subsection{Comparison with state-of-the-art models}
Table~\ref{Tab:result1} shows the results obtained by \texttt{PREDICT-AI} and baseline DRP models on three distinct drugs (5-Fluorouracil, Cisplatin, and Paclitaxel), with more than 80 (sample, drug) pairs in the TCGA dataset and using the same train-test splits in~\cite{jayagopal2023multi}. \texttt{PREDICT-AI} surpasses baselines in overall prediction performance, as well as on two other drugs. 
\texttt{PREDICT-AI} achieves the highest overall evaluation scores, with 64.96$\%$ for AUROC and 84.85$\%$ for AUPRC, outperforming the second-highest scores achieved by DruID (62.36$\%$ for AUROC and 82.06$\%$ for AUPRC). On Cisplatin and 5-Fluorouracil, \texttt{PREDICT-AI} achieves an improvement of 4.99\% and 6.57\% for AUROC, and 3.96\% and 0.72\% for AUPRC compared to the second-best performing method.
The superiority of \texttt{PREDICT-AI} in DRP over baseline models demonstrates the advantages of our transformer-based approach of modeling input sequences and use of auxiliary survival information.


\subsection{Ablation Tests}
\begin{figure*}[t]
\centering
\includegraphics[width=1.95\columnwidth]{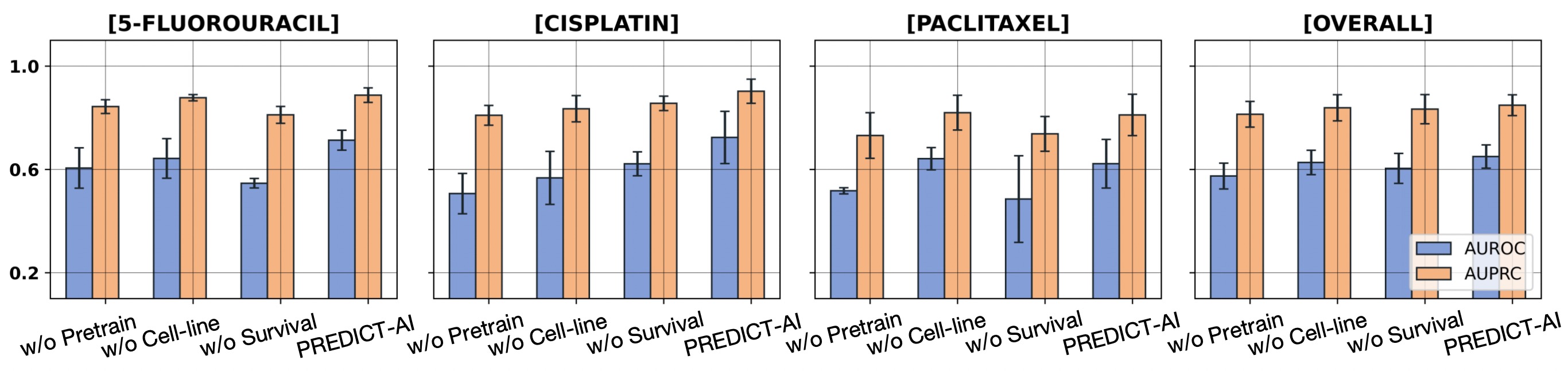}
\caption{Effects of individual components in the \texttt{PREDICT-AI} model.}
\label{fig02}
\end{figure*}

To evaluate the effectiveness of our proposed model, we conduct an ablation study to examine three specific components in \texttt{PREDICT-AI}: one without the pre-training/\texttt{TransformerMTLR} component (`\textit{w/o Pretrain}'), another without utilizing the cell-line/AUDRC predictor (`\textit{w/o Cell-line}'), and a third without the survival predictor in \texttt{TransformerDRP} (`\textit{w/o Survival}'). 
Figure~\ref{fig02} illustrates that the pretrained \texttt{TransformerMTLR} model contributes to enhancing the DRP outcomes, e.g., the AUROC and AUPRC achieved by \texttt{PREDICT-AI} without pretraining are 0.5774 and 0.8135, respectively, significantly lower than the metrics obtained by \texttt{PREDICT-AI}. 
Besides, incorporating cell-line data for the AUDRC predictor and reintroducing the survival predictor in \texttt{TransformerDRP} also contribute to the DRP performance.

\begin{table*}[htbp]
\caption{Experimental results comparing PREDICT-AI against extant DRP methods on 3 distinct drugs in TCGA patients. Best scores are in bold, second-best scores are underlined. ($\%$).}\label{Tab:result1}
\centering
\small
\renewcommand\arraystretch{1.0}
\setlength{\tabcolsep}{0.95mm}{
\begin{tabular}{c|cc|cc|cc|cc}
\toprule
  & \multicolumn{2}{c|}{\textbf{5-Fluorouracil}} & \multicolumn{2}{c|}{\textbf{Cisplatin}} &  \multicolumn{2}{c|}{\textbf{Paclitaxel}} & \multicolumn{2}{c}{\textbf{Overall}} \\
  \cline{2-9}
  & AUROC & AUPRC & AUROC & AUPRC & AUROC & AUPRC & AUROC & AUPRC \\
\midrule
  \textbf{CODE-AE} & 61.77($\pm$11.63) & \underline{88.02($\pm$4.85)} & 38.09($\pm$16.34) & 71.88($\pm$8.46) & 44.83($\pm$15.08) & 74.68($\pm$5.11) & 47.15($\pm$7.82) & 73.74($\pm$6.90) \\
  \textbf{TCRP} & 48.39($\pm$11.96) & 77.87($\pm$9.49) & 50.86($\pm$16.75) & 79.60($\pm$10.73) & 60.66($\pm$24.22) & 78.69($\pm$16.73) & 47.36($\pm$3.26) & 75.70($\pm$4.03) \\
  \textbf{TUGDA} & 58.39($\pm$16.94) & 83.40($\pm$7.69) & 37.47($\pm$4.36) & 72.07($\pm$3.75) & 41.13($\pm$18.28) & 71.67($\pm$1.03) & 46.18($\pm$3.18) & 75.42($\pm$2.22) \\
  \textbf{Velodrome} & 50.91($\pm$19.54) & 79.08($\pm$15.04) & 48.61($\pm$9.73) & 78.80($\pm$7.97) & \underline{63.26($\pm$26.45)} & 80.02($\pm$16.34) & 52.56($\pm$4.93) & 77.62($\pm$0.48) \\
  \textbf{DruID} & \underline{64.73($\pm$8.73)} & 85.55($\pm$6.43) & \underline{67.38($\pm$10.63)} & \underline{86.30($\pm$7.35)} & \textbf{63.43($\pm$4.97)} & \textbf{82.55($\pm$6.83)} & \underline{62.36($\pm$3.60)} & \underline{82.06($\pm$5.02)} \\
\midrule
  \textbf{PREDICT-AI} & \textbf{71.30($\pm$3.87)} & \textbf{88.74($\pm$2.82)} & \textbf{72.37($\pm$10.10)} & \textbf{90.26} ($\pm$4.68) & 62.19($\pm$9.42) & \underline{81.08($\pm$8.02)} & \textbf{64.96($\pm$4.50)} & \textbf{84.85($\pm$4.02)} \\
\bottomrule
\end{tabular}}
\end{table*}

\section{Deployment in a Clinical Setting}

Our Treatment Recommendation System has been deployed at \href{https://pharmacope.ai/}{https://pharmacope.ai/}.
The backend architecture, described in Appendix \ref{backend}, processes the input PDF diagnostic panel reports, extracts the mutations therein, runs a DRP model and returns a list of 10 drugs with the best predicted RECIST probability scores along with supporting evidence to the user interface.

\subsection{User Interface}
\label{sec: frontend}

\begin{figure}[h!]
  \centering
  \includegraphics[width=\linewidth]{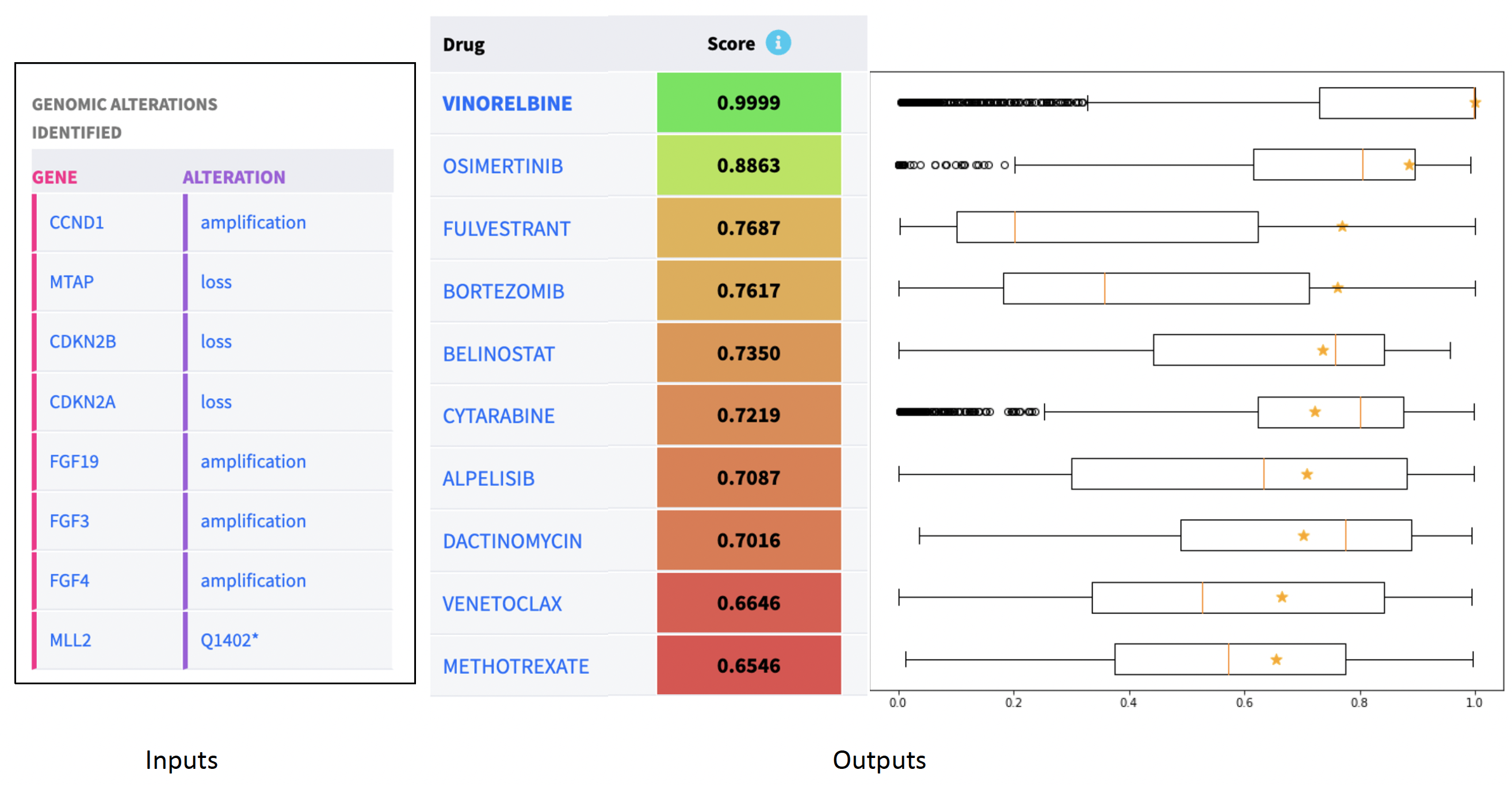}
  \caption{Left panel shows mutations present in patient genomic profile. Middle and right panels displays top 10 recommendations with supporting evidence as boxplots.}
  \label{fig_recommendations}
\end{figure}


The interface, used by clinicians, shows the input set of mutations in the patient genomic profile (Appendix Figure \ref{fig_inputs_trs}),  the top 10 drug recommendations with their predicted scores, 
and supporting evidence (Figure \ref{fig_recommendations}). 
As discussed earlier, in addition to these predictions, the Molecular Tumor Board (MTB) requires multiple available sources of information, which may be incomplete and indirect, to arrive at the final treatment decision.
Based on extensive discussion among us -- computer scientists and oncologists -- we arrived at the following sources of information which have been useful in the clinical decision-making process.
\begin{itemize}[noitemsep,topsep=0pt,labelindent=0em,leftmargin=*]
    \item Model explainability: 
    Many Explainable AI (XAI) algorithms can be used to provide model explainability~\cite{dwivedi2023explainable}. E.g., feature importance scores or attention scores can be used to highlight important genes. While clinicians are aware that the predictions are a result of complex series of computations, and XAI scores may not be a faithful or complete representation of the model's inner working, these inputs may become useful starting points for discussing potential targeted interventions.
    \item Auxiliary drug databases: Several publicly available portals such as Pandrugs~\cite{pineiro2018pandrugs}, collect information from multiple databases to
    display known information on drug-gene associations. Such information can be used to assess the available level of evidence and 
    examine the biological plausibility of the drugs recommended, based on patient cancer type and known drug targets. 
    \item Drug-level validation across patients: 
    For each predicted drug, clinicians are interested in knowing the difference in the prediction (for the input patient) to the predictions on a reference (or validation) set of patients. 
    The reference set is often chosen as a cohort of patients of the same cancer type. Note that it is unlabelled, i.e., without drug response or survival labels.
    E.g., in Fig. \ref{fig_recommendations}, we see the distribution of predictions for each drug on a reference dataset and the prediction on the input patient overlaid with an asterisk. If the input patient's prediction is an outlier (e.g., for Fluvestrant), indicating an unusually good prediction vis-a-vis the reference cohort, it highlights a drug that could be considered further. This may be quantified with
    robust z-scores by subtracting the median predicted score on the validation dataset and dividing by the interquartile range.
    \item Patient-level validation across drugs: 
    Lack of sufficient labelled data during model training can result in, for some input patients, very similar scores on all input drugs. In such cases, our confidence in the predictions should be low.
    Hence, it is helpful to rule out such cases by examining the distribution of predicted scores of the patient, across all drugs, not just the top 10, as shown in Fig. \ref{fig_swarmplot}. This can also be quantified using z-scores. 
\end{itemize}
\subsection{Clinical Treatment Planning Process}
\label{sec:cdss_process}
Evaluating the drugs predicted by a DRP model in the face of insufficient clinical validation is a non-trivial and challenging problem.
Retrospective evaluation on clinical datasets (e.g., on TCGA, Table \ref{Tab:result1}), remains limited since the number of possible combinations of mutations is exponentially large.
Hence, such evaluation, although a necessary first step, is insufficient to ensure that the predicted drug will be effective when used on an unseen patient.
A clinical trial of the DRP model itself can be done, which also would yield limited data due to the same reasons. Moreover, during such a trial, the choice of drug to be administered to an enrolled patient is not straightforward -- a problem which will also arise when the system is deployed for routine clinical use. 

A clinical trial is ongoing for our treatment recommendation system at the National University Hospital, Singapore~\cite{trialID}. 
Before commencing the trial, an important validation for our DRP model was the presence of cancer type-specific standard-of-care drugs among the top predicted drugs on test patients, 
e.g., 5-Fluorouracil, Oxaliplatin, Irinotecan in colorectal cancer, and Cisplatin, Paclitaxel in ovarian cancer.
This led to clinicians gaining confidence in our model. These drugs may not always be the topmost in the recommended list, as the model learns from only mutations and drug response in cell lines and patient data. Thus, it was important to display higher number of recommendations and we chose 10.




The decision of the most appropriate drug for a patient, is taken in a collective manner involving all members of the MTB.
An important lesson learnt before and during the trial was that many additional sources of information are required by the MTB to make the final treatment decision.
The top 10 recommendations and supporting evidence (Section \ref{sec: frontend}) from various sources are used together for decision making. 
Further, patient-specific details are considered such as possibility of side-effects, avoiding 
drugs that caused a disease progression in earlier treatment cycles. 



\section{Conclusion}
We design a new cancer drug response prediction (DRP) model, \texttt{PREDICT-AI}, which predicts the efficacy of a given drug on a patient's mutation profile from clinical diagnostic panels.
\texttt{PREDICT-AI} faithfully models the sequential structure of input mutations using transformers. Further, it effectively utilizes auxiliary patient survival information for training.
These modeling steps address limitations of extant models and lead to improved performance on benchmark data compared to state-of-the-art alternatives.

We present the architecture of a treatment recommendation system (TRS) which internally uses DRP models.
The TRS is deployed in a cancer-specific clinical setting -- a Molecular Tumour Board (MTB) whose members collectively makes treatment decisions.
We discuss the technical and clinical considerations required to make treatment decisions in this challenging scenario where validation of the predictions from any DRP model is difficult to obtain. In particular, we present various forms of indirect supporting evidence which were found to be useful in building clinician's trust in the model as well as in making the final choice of a suitable drug.

Our work opens several avenues for future research.
Other modeling approaches that build on PREDICT-AI may be investigated, e.g., other forms of tokenization, survival analysis and multi-task learning.
The use of additional unlabelled data during pre-training and various forms of attention-based explainability can be explored.
A limitation of our work, which can be addressed in the future, is that it does not model rearrangement-based alterations which are also available in diagnostic panels. 
Finally, more work can be done on designing methods and systems that can integrate information from DRP models and auxiliary sources of information to effectively faciliate collective decision-making in MTBs. 



\bibliographystyle{ACM-Reference-Format}
\bibliography{references}

\appendix
\section{Supplementary Material}
\subsection{Data Processing}
\label{sec:data_processing}
In cell lines, we consider only labelled samples with a documented AUDRC score, resulting in 689 samples. In the TCGA dataset, we only consider the samples belonging to the following projects/cancer types - LUAD, STAD, HNSC, SKCM, BLCA, UCEC, COAD, LUSC, BRCA, CESC. We also restrict our analysis to a set of drugs (Cisplatin, Paclitaxel, 5-Fluorouracil, Gemcitabine, Docetaxel, Cyclophosphamide) administered to more than 50 TCGA samples. The resulting set of 470 TCGA patients have a documented RECIST response. In the GENIE BPC CRC dataset, we extract the progression-free survival duration as the time between the start of the first line of treatment and the subsequent progression, according to the medical oncologist's assessment. We obtain this information for 5-Fluorouracil, resulting in 822 CRC samples with documented PFS values. A similar analysis on the GENIE BPC NSCLC v2.0-public dataset resulted in 900 patients (i.e. 1490 samples) with documented PFS values for 16 drugs. 

\subsubsection{Annotation using ClinVar, GPD and Annovar}
For each sub-gene-level mutation, we follow the variant annotation procedure described in~\cite{jayagopal2023multi}. However, we do not perform the gene-level aggregation step. This results in a 17 dimensional binary vector from Annovar~\cite{wang2010annovar}, a 3-dimensional binary vector each from GPD~\cite{li2020protein} and ClinVar~\cite{landrum2018clinvar}, together resulting in a 23-dimensional vector per mutation. As in~\cite{jayagopal2023multi}, we only use mutations and do not use the copy number variation information available in these datasets.

\label{algorithm}
{\centering
\begin{minipage}{.975\linewidth}
\removelatexerror
\begin{algorithm*}
\SetAlgoLined
\KwData{Survival data $\mathcal{P}_{CRC}$, $\mathcal{P}_{NSCLC}$; TCGA patient samples $\mathcal{P}_{TCGA}$; cell line data $\mathcal{P}_{CL}$; initialization parameters;}
\KwResult{Drug response prediction results $\hat{Y}$.}
// \texttt{Module 1: \texttt{TransformerMTLR}} \\
S $\leftarrow$ $\{t_i\}_{1}^{N}$ \ \ \ //\ \texttt{Framing the multi-task learning problem} \\
\For{$e\in \mathrm{epochs}$}{
\For{$b\in$ $\mathrm{Batches}$ $\ in\ \{\mathcal{P}_{NSCLC},\mathcal{P}_{CRC}\}$}{
$\bar{X} \leftarrow \mathrm{transformer}(X_G||X_M)$ \ \ //\ \texttt{Learning genes and mutation representations using transformer} \\
$\bar{X}_D \leftarrow \mathrm{MLP_{Drug}}(X_D)$ \ //\ \texttt{Learning drug representations using MLP} \\
$\hat{S} \leftarrow \mathrm{MLP}(\bar{X}||\bar{X}_D)$ \ \ //\ \texttt{MTLR predictor} \\
${\mathcal{L}_S}$ $\leftarrow$ Equation (2) \ // \texttt{Loss for survival prediction}
}
}

// \texttt{Module 2: \texttt{TransformerDRP}} \\
$\mathrm{transformer}$, $\mathrm{MLP_{Drug}}$ $\leftarrow$ Load(*) \ //\ \texttt{Pretrained models} \\
\For{$e\in \mathrm{epochs}$}{
\For{$b\in \mathrm{Batches}$ in $\{\mathcal{P}_{NSCLC},\mathcal{P}_{TCGA},\mathcal{P}_{CL}\}$}{
${\mathcal{L}_S}$ $\leftarrow$ \texttt{TransformerMTLR}($\mathcal{P}_{NSCLC}$)\ //\ \texttt{Survival Prediction}\\
//\ \texttt{AUDRC Prediction: $\mathcal{P}_{CL}$}\\
$\bar{X},\bar{X}_D \leftarrow \mathrm{transformer}(X_G||X_M),\mathrm{MLP_{Drug}}(X_D)$ \\
$\hat{Y} \leftarrow \mathrm{MLP}(\bar{X}||\bar{X}_D)$ \\
${\mathcal{L}_C}$ $\leftarrow$ Equation (8) \ // \texttt{Loss for AUDRC prediction}

//\ \texttt{RECIST Prediction: $\mathcal{P}_{TCGA}$}\\
$\bar{X},\bar{X}_D \leftarrow \mathrm{transformer}(X_G||X_M),\mathrm{MLP_{Drug}}(X_D)$ \\
$\hat{R} \leftarrow \mathrm{MLP}(\bar{X}||\bar{X}_D)$ \\
${\mathcal{L}_R}$ $\leftarrow$ Equation (7) \ // \texttt{Loss for RECIST prediction}
}
$\mathcal{L}$ $\leftarrow$ $\mathcal{L}_S$ + $\mathcal{L}_R$ + $\mathcal{L}_C$ \ // \texttt{Jointly train 3 loss functions}
}
\caption{The \texttt{PREDICT-AI} algorithm.}
\end{algorithm*}
\end{minipage}
\par
}

\subsection{Backend Architecture}
\label{backend}
\begin{figure}[h!]
  \centering
  \includegraphics[width=\linewidth]{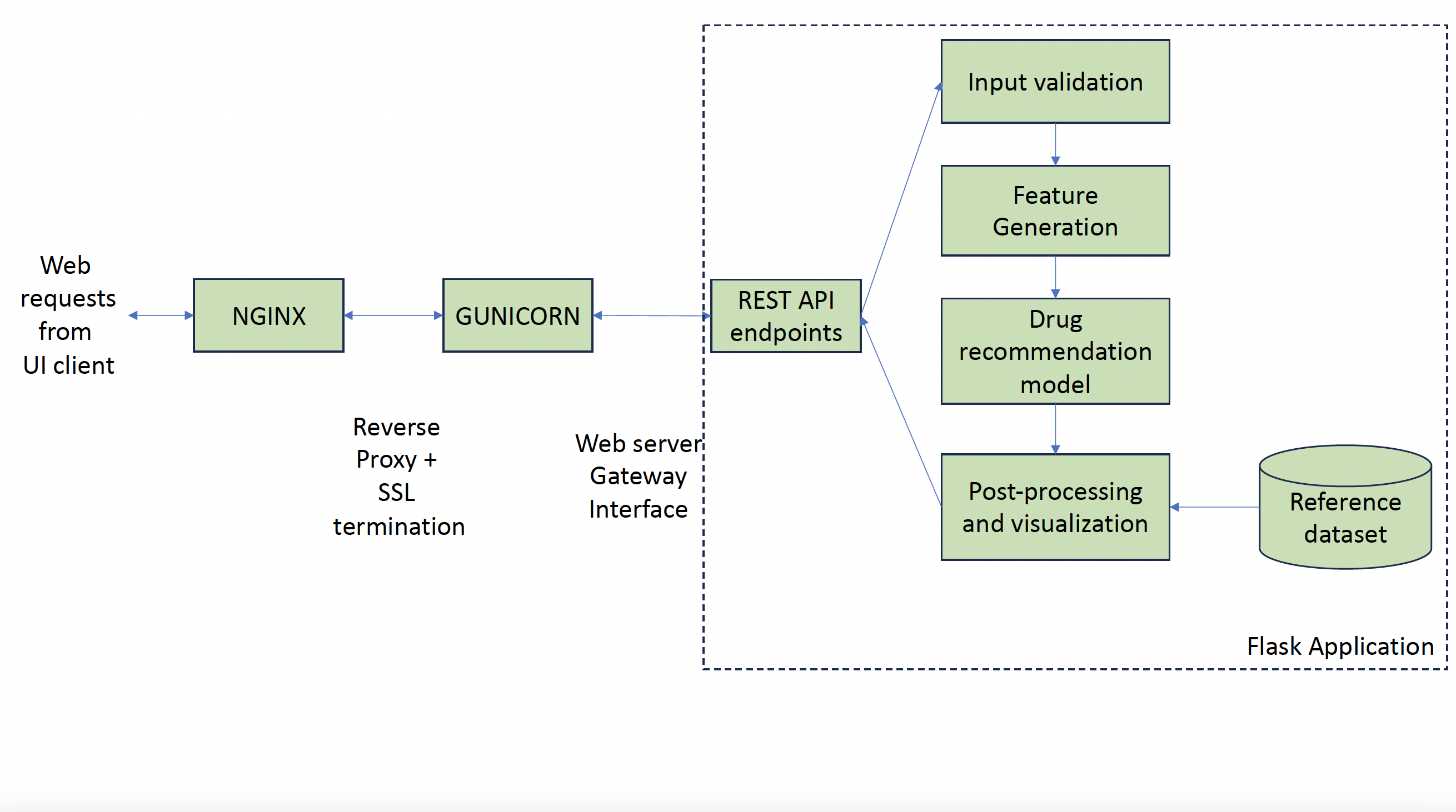}
  \caption{Backend Architecture of TRS}
  \label{fig_deployment}
\end{figure}

Here, we describe the backend architecture (Figure \ref{fig_deployment}) of the TRS used to deploy a 
DRP model. Our deployment has been set up on an Amazon LightSail instance, which serves requests from web clients. These requests are handled by an NGINX reverse proxy, which also does SSL termination. Calls from the user interface (described in Section \ref{sec: frontend}) are made via API calls served by a Flask application. This application also uses a \href{https://gunicorn.org/}{Gunicorn} HTTP server. The API call to obtain recommendations is invoked with the set of patient mutations provided as inputs.
FoundationOne reports may also be directly used as inputs -- the PDF is parsed to extract out mutation information.
The input mutations are first validated to ensure proper formatting, followed by feature generation (annotations or other strategies based on the DRP model inputs). These are then passed through the DRP model for inference on a set of 70 drugs approved for clinical use in Singapore.
Drugs with top 10 predicted scores are provided along with supporting evidence (as described in \ref{sec: frontend}) to the user interface. Minimal unit-testing of the API endpoints is achieved through \textit{pytest}. The deployment can be accessed at \href{https://pharmacope.ai/}{https://pharmacope.ai/}.

\subsection{Implementation}
To reproduce the experimental results outlined in our paper, we specify the parameter ranges used. The representation dimension is empirically fixed at 64. The learning rate varies from 0.0001 to 0.05. Epochs range from 100 to 500, with a default dropout value of 0.1. Batch sizes are selected from the range [128, 256, 512]. Our model offers two optional optimizers: Adam and SGD. The default number of attention heads is set to 8.
All codes, data and experimental settings of the model are released at \href{https://github.com/CDAL-SOC/PREDICT-AI}{https://github.com/CDAL-SOC/PREDICT-AI}.

\subsection{Supplementary Figures}
All icons in this manuscript have been generated using Flaticon.


\begin{figure}[h]
    \centering
    \includegraphics[width=\linewidth]{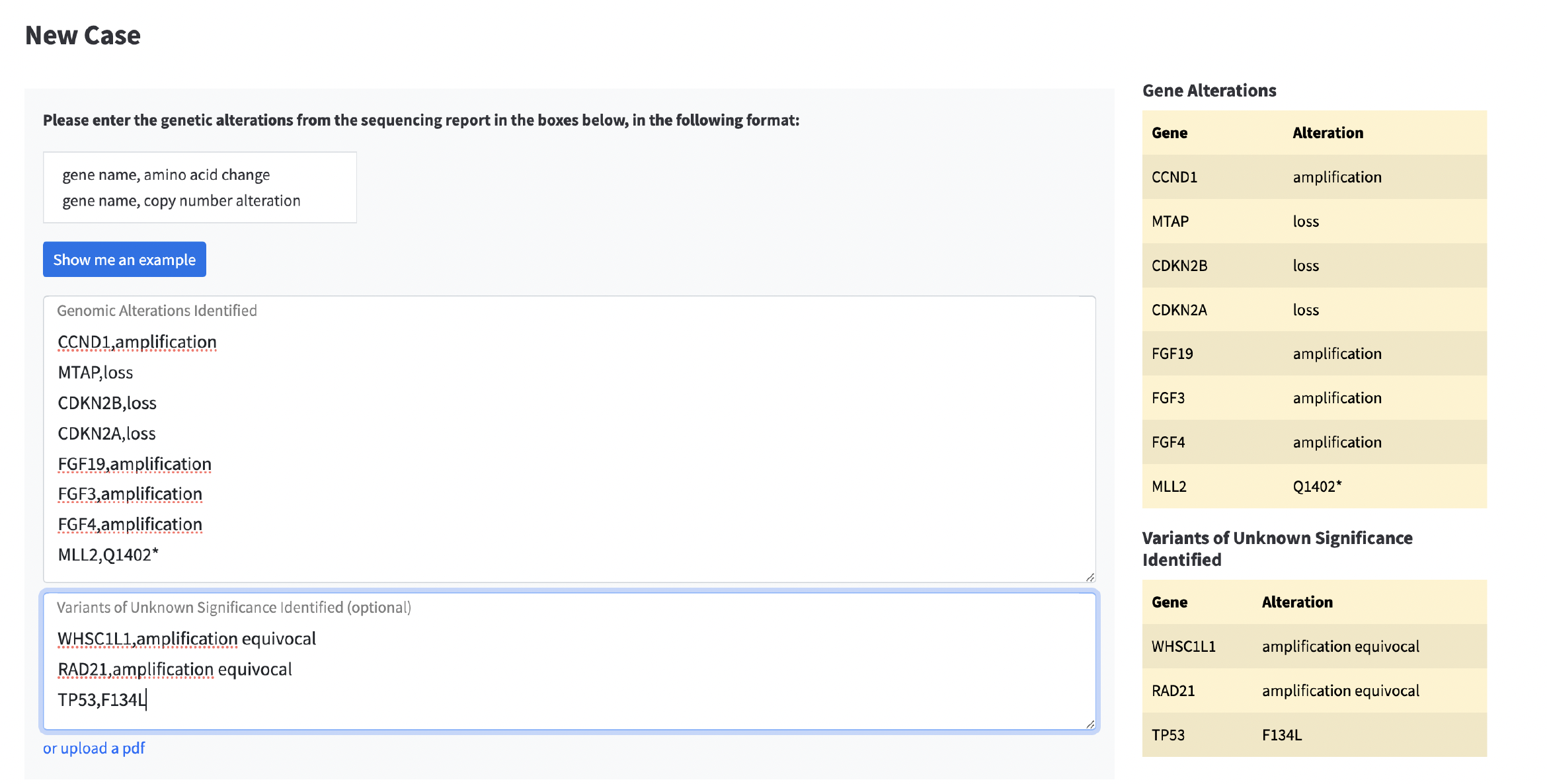}
    \caption{Patient genomic profile fed as input to TRS.}
    \label{fig_inputs_trs}
\end{figure}

\begin{figure}
    \centering
    \includegraphics[width=\linewidth]{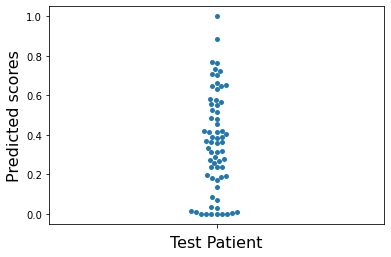}
    \caption{Swarmplot indicating predicted scores across all approved drugs for a test patient.}
    \label{fig_swarmplot}
\end{figure}

\subsection{Additional Baseline Models Comparison}
We compare the test concordance indices of baseline models using binary (BIN324) and annotated input on CRC dataset, divided into train, validation, test segmentation (64:16:20). The results show that compared with the binary input, the mutation auxiliary information has no obvious strengthening effect on the model.

\begin{table}[h]
   \caption{Comparison of binary BIN324 and annotated inputs on baseline models for survival prediction of CRC dataset.}
  \label{tab:surv_results_2}
   \begin{tabular}{c|cc}
    \toprule
    Model & BIN324 & Annotated \\
    \midrule
    DeepSurv & 0.4747 & 0.5127 \\
    DeepHit & 0.5307 & 0.4659 \\
    LogisticHazard & 0.5069 & 0.5251 \\
    MTLR & 0.5061 & 0.4722 \\
    PCHazard & 0.5410 & 0.4895 \\ 
    PMF & 0.5027 & 0.4697 \\
    \bottomrule
  \end{tabular}
\end{table}




\end{document}